\documentclass[journal]{IEEEtran}

\usepackage{cite}

\usepackage[pdftex]{graphicx}
\usepackage[section]{placeins}

\usepackage{makecell}

\usepackage{amsmath,amssymb}

\usepackage[caption=false]{subfig}
\usepackage{enumerate}

\begin{document}

\title{Clustering Introductory Computer Science Exercises Using Topic Modeling Methods}

\author{Laura~O.~Moraes and Carlos~Eduardo~Pedreira,~\IEEEmembership{Senior Member,~IEEE}% <-this % stops a space
\thanks{Manuscript received 2019. The work of Laura O. Moraes was supported by the Brazilian National Research Council (CNPq) under Grant 141089/2016-4. The work of Carlos Eduardo Pedreira was supported in part by the Research Foundation of the State of Rio de Janeiro (FAPERJ) under Grant E26/202.838/2017-CNE, in part by the Coordination of Improvement of Higher Education Personnel (CAPES) under Grant PROEX - 1201036, and in part by the Brazilian National Research Council (CNPq) under Grant 306258/2019-6. \textit{(Corresponding author: Laura O. Moraes.)}}%
\thanks{The authors are with the Systems and Computing Engineering Department (COPPE-PESC), Universidade Federal do Rio de Janeiro (UFRJ), Rio de Janeiro, 21941-914, Brazil (e-mail: lmoraes@cos.ufrj.br; pedreira56@gmail.com).}}

% The paper headers
\markboth{\copyright 2021 IEEE, Accepted article. Published in IEEE Transactions on Learning Technologies (doi: 10.1109/TLT.2021.3056907)}%
{Moraes \MakeLowercase{\textit{et al.}}: Clustering Introductory Computer Science Exercises Using Topic Modeling Techniques}

% make the title area
\maketitle

\begin{abstract}
\textbf{This  is  the  accepted  version,  to  read  the  final  version  published  in  2021  in  the  IEEE Transactions on Learning Technologies  (IEEE TLT), please go to: http://doi.org/10.1109/TLT.2021.3056907. Personal use of this material is permitted.  Permission from IEEE must be obtained for all other uses, in any current or future media, including reprinting/republishing this material for advertising or promotional purposes, creating new collective works, for resale or redistribution to servers or lists, or reuse of any copyrighted component of this work in other works.} 

Manually determining concepts present in a group of questions is a challenging and time-consuming process. However, the process is an essential step while modeling a virtual learning environment since a mapping between concepts and questions using mastery level assessment and recommendation engines are required. We investigated unsupervised semantic models (known as topic modeling techniques) to assist computer science teachers in this task and propose a method to transform Computer Science 1 teacher-provided code solutions into representative text documents, including the code structure information. By applying non-negative matrix factorization and latent Dirichlet allocation techniques, we extract the underlying relationship between questions and validate the results using an external dataset. We consider the interpretability of the learned concepts using 14 university professors' data, and the results confirm six semantically coherent clusters using the current dataset. Moreover, the six topics comprise the main concepts present in the test dataset, achieving 0.75 in the normalized pointwise mutual information metric. The metric correlates with human ratings, making the proposed method useful and providing semantics for large amounts of unannotated code.
\end{abstract}


\section{Introduction}\label{sec:introduction}

\IEEEPARstart{M}{easuring} student's knowledge requires an understanding of which educational concepts are needed
to answer each question.
%an adequate level of education and understanding to answer their individual questions.
Recently, open online courses and intelligent tutoring systems are widely adopted learning environments. Their popularity increases the demand for tools to map questions to concepts correctly since students' mastery level assessment and next steps recommendation depend on these mappings.

However, manually identifying the concepts required to answer questions can be time-consuming and difficult, increasing the need for tools to assist teachers in the tasks. Desmarais~\cite{desmarais_mapping_2012} suggested that even partial automation of the process can be highly desirable. 
Besides decreasing the manual labeling required from the experts, the process automation also results in a more objective and replicable mapping. Applying supervised machine learning-based solutions is not entirely appropriate because it requires a considerable amount of labeled data. Also, a question can relate to multiple concepts, increasing the complexity of the labeling task. The traditional unsupervised learning methods, such as K-means and hierarchical clustering, are also not suitable for this task because it is hard to determine each cluster's features. 

This paper proposes unsupervised semantic methods, known as topic modeling techniques~\cite{Lee1999, Blei.DavidM.2003, Steyvers2010, hofmann_probabilistic_2013}, as more interpretable methods for experts, to be applied in introductory computer science problems. Specifically, we propose modeling code snippets as text documents and use topic modeling techniques to extract and improve the semantic relationships between them, providing technology to support concept identification experts.

Our key research questions are summarized as follows: 1) how can semantic relationships be extracted and structured from code? 2) how can humans read, interpret, and use the extracted relationships? 

The main contributions of this paper include:
\begin{enumerate}[(i)]
    \item A tokenization structure to transform raw code snippets into a document-term matrix.
    \item A code-clustering method to optimize positively correlated metrics for human-interpretability.
    \item Experts validation, illustrating how the proposed method can support questions' exercise labeling using each topic's terms.
\end{enumerate}

The proposed code-clustering pipeline builds a document-term matrix with a code tokenizer by comparing various methods, including clustering algorithms. We compare non-negative matrix factorization (NMF)~\cite{Lee1999}, latent Dirichlet allocation (LDA)~\cite{Blei.DavidM.2003}, and K-means~\cite{lloyd_least_1982}, as a baseline. The models were evaluated with the UMass and UCI coherence metrics~\cite{newman_automatic_2010, mimno_optimizing_nodate, aletras_evaluating_nodate, Roder2015} using the top-5 and top-10 terms. According to the metrics scores, we selected the two best-ranked models using Fagin’s algorithm~\cite{fagin1996combining}. The LDA-based clustering approach provides the most interpretable results from these models. Fourteen professors manually contextualize the LDA-based clustering in the Computer Science 1 (CS1) domain, demonstrating how the proposed method could be used to facilitate the clusters' interpretability.

The next section begins by reviewing the manual, supervised, and unsupervised concept identification approaches. Code-clustering techniques inside the educational data mining (EDM) and software engineering contexts are reviewed together with existing LDA proposals to handle short texts. We describe a proposed method to cluster code in Section~\ref{sec:methods}. The main challenges facing the CS1 context are the answers' small size and document term-matrix sparsity. Our proposed code tokenizer overcame these problems using code structure information to augment the corpus. The results are shown in Section~\ref{sec:results}, where the best two clustering schemes are analyzed based on the coherence evaluation metrics. This section also demonstrates how professors can get an overview of the required concepts from these results. Finally, Section~\ref{sec:conclusion} presents the conclusion and future work directions.

\section{Related Work}
The existing methods to identify concepts from a set of CS1 exercises involve manual work and input from experts~\cite{Sheard2011, Petersen2011}. For example, Sheard \textit{et al.}~\cite{Sheard2011} characterized introductory programming examination questions according to their concept areas, question style, and required skills. Participants manually classified the questions and the determined topics covered alongside the necessary skill levels to solve them. Nonetheless, applying a successful approach in a different set of exercises requires a new manual labeling stage, which may not be achievable.

One strategy to overcome this issue and minimize the domain experts' workload is to apply supervised learning. Previous research in question classifications used supervised learning to classify questions according to the level of difficulty~\cite{Fei2003}, Bloom's taxonomy~\cite{Supraja2017}, answer type~\cite{Godea2018}, and subject~\cite{Gonzalez}. In Godea \textit{et al.}~\cite{Godea2018}, the features are derived from the questions, using part-of-speech tags, word embeddings, inter-class correlations, and manual annotation. Supraja \textit{et al.}~\cite{Supraja2017} use a grid search to analyze different combinations of weight schemes and methods to find the best set of parameters to build a supervised model to classify questions given Bloom's Taxonomy. Its main cost is the manual annotation of all labels, impractical when applying to large datasets. Unsupervised learning can group similar items without a predefined label, but it is harder to ascertain the results since there is no objective goal to analyze, and evaluating the clustering outcomes becomes a subjective task. Unsupervised learning techniques have been used to address EDM problems~\cite{nunn_learning_2016, Dutt2017, rodrigues_educational_2018, hernandez-blanco_systematic_2019, aldowah_educational_2019}. For example, Trivedi \textit{et al.}~\cite{trivedi_spectral_2011} use spectral clustering with linear regression to predict student performance. In the questions' classification context, an unsupervised approach using K-means, as a clustering algorithm~\cite{DeBarrosBorgesReisFigueira2008}, was proposed to group similar learning objects (such as handouts, exercises, complementary readings, and suggested activities). Still, K-means does not provide a list of features that best characterize each cluster, making the expert infer them manually by reading a sample of each cluster's exercises.

It is possible to use the code provided as answers to the exercises as input while restricting the concept identification to the CS1 domain. Unsupervised learning using code as input is achieved by calculating their abstract syntax trees (ASTs) and clustering the most similar ones; besides, various strategies calculate similarity among trees. Huang \textit{et al.}~\cite{Huang2013} and Nguyen  \textit{et al.}~\cite{Nguyen2014} use edit tree distance, Paasen \textit{et al.}~\cite{Paassen2016} apply sequential clustering algorithms, and Price \textit{et al.}~\cite{Price2017} and Mokbel \textit{et al.}~\cite{Mokbel2013} fragment each AST into subgraphs, pair similar subgraphs and compute the distance between the subgraphs within a pair. Concepts identification in these scenarios requires experts to read samples of exercises from each cluster to infer the relevant features. It is also unclear how to generalize it to other domains where the answers are not constructed directly using a tree or a graph.

We propose a CS1 concept identification method to support experts in labeling concepts by providing the relevant features in each cluster. This method could potentially be extended to different domains since it only requires text inputs. We develop a code tokenizer that uses its code structure to augment and improve the corpus. We apply topic modeling techniques where the documents are the code provided as answers to the exercises. In the software engineering field, concepts are extracted from code using programmers' annotations~\cite{yihan_annotation2018}. Similarly, code clustering, identifying similar features/concepts in a repository, predicting bugs, analyzing/predicting source code evolution, tracing links between modules, and detecting clones are widely applied in the same field~\cite{Chen2016}. For the EDM, Azcona \textit{et al.}~\cite{azcona_code2vec2019} created a Python code submission tokenizer to setup features to generate code embeddings. Unlike the software engineering context, code snippets in CS1 are small in size and lack annotations.

Although the LDA in Blei \textit{et al.}~\cite{Blei.DavidM.2003} is a common technique in topic modeling, it does not perform well in short texts (code in this context) because the traditional way of extracting terms does not provide enough textual words to characterize a specific topic~\cite{chen_short_2016, abolhassani_extracting_2019}. It is necessary to decrease the latent document-topic or word-topic spaces, making them more specific for each context. Hsiao \textit{et al.}~\cite{hsiao_topic_2015, hsiao_enriching_2017} propose a topic-facet LDA model using sentence LDA (SLDA) with a facet representing a more specific topic and all words from a sentence belonging to the same facet. Zhao \textit{et al.}~\cite{zhao_comparing_2011} decrease the latent space by creating a common word distribution with denominated background words, which are the same for every topic. 
%In this way, a topic has a smaller specialized word-topic latent space.
Steyers \textit{et al.}~\cite{steyvers_probabilistic_2004} and Rosen-Zvi \textit{et al.}~\cite{rosen-zvi_learning_2010} adopted a similar strategy. In their method, the generative process to create a document decreases the space by choosing an author and then choosing a topic. Li \textit{\textit{et al.}}~\cite{li_tag-weighted_nodate} use a distribution over tags to restrict the latent topic space before inferring the documents' topic distribution.

Another approach to overcome the lack of textual words is to increase the representation, so Weng \textit{et al.}~\cite{weng_twitterrank_2010} proposed an LDA to cluster subjects' twitterers (people who tweet). Their method increases document representation by aggregating all tweets from a single user into one document. Abolhassani and Ramaswamy~\cite{abolhassani_extracting_2019} enhance short semi-structured texts by augmenting the corpus' structure, which is similar to the method we adopted in this paper. In our proposed tokenizer, we increase the vocabulary size (total terms) from 287 to 2388 and the average of terms per document from 23 to 137. We maintain a 95\% sparsity, which agrees with the sparsity of the long-text documents from Syed and Spruit~\cite{syed_full-text_2017} and Zhao \textit{et al.}~\cite{zhao_comparing_2011}, i.e., successfully clustered using topic modeling techniques.

\section{Methods}
\label{sec:methods}
The methodology is categorized into three main tasks: 1) we generate a database by crawling different Python web tutorials, 2) run code-clustering experiments to group exercises into topics, and 3) ask experts to contextualize the clusters into CS1 concepts. Task 1 prepares the data to investigate the research questions. Task 2 explores ways of extracting semantic relationships from code [research question 1 (RQ1)] by proposing a code tokenizer and comparing various data transformation methods and topic modeling algorithms. In task 3, we ask professors to read and interpret task 2, giving support for RQ2. All the scripts used in this study analyses were achieved using Python and open-source Python libraries. The code is available at https://github.com/laura-moraes/machine-teaching.

%%%%%%%%%%%%%%%%%%%% Table No: 1 ends here %%%%%%%%%%%%%%%%%%%%

\subsection{Dataset}
The objective of the experiment is to find semantically related CS1 code solutions written in Python. We chose four introductory online tutorials: Practice Python~\cite{Michele_Pratusevich_2017}, Python School~\cite{Sue_Sentance__and__Adam_McNicol_2016}, Python Programming Exercises~\cite{Jeffrey_Hu_2018}, and W3Resource~\cite{W3Resource_2018} that provide both solutions and exercise statements. Since the sources do not have label topics or follow a course curriculum with structured syllabus topics, we work in an unsupervised environment. We crawled 54 exercises for the training set. The code snippets are functions with an average of 9 lines/code.

For the test dataset, we collected solutions from another set of exercises given to us by the CS1 professors at the Federal University of Rio de Janeiro (UFRJ). There are 65 different problems with their respective solutions in the dataset. As the training set, the code snippets are functions with an average of 7 lines/code.

\subsection{Code-Clustering Pipeline}
\label{ssec:pipeline}
The code-clustering pipeline takes as input Python code snippets, which are semi-structured text documents. By using topic modeling techniques, the pipeline outputs an underlying structure within the semi-structured corpus. It contains the topics present in the code snippets and the most relevant words that characterize them. This paper is based on the assumption that code snippets with similar CS1 concepts share identical terms. Therefore, based on this assumption, the extracted topic underlying structure can be interpreted as CS1 concepts or groups of CS1 concepts present in the code snippets. 

\begin{figure*}[h!]
% \begin{figure}[h!]
    \centering
    \includegraphics[width=0.8\textwidth]{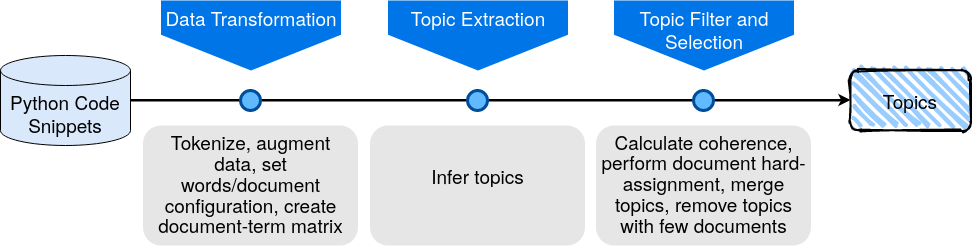}
	\caption{Code-clustering pipeline overview~\cite{abolhassani_extracting_2019}.}
	%Adapted from Abolhassani \textit{et al.}~\cite{abolhassani_extracting_2019}.}
	\label{fig:pipeline}
\end{figure*}

The code-clustering pipeline starts by transforming the original data to the proper format expected by the topic modeling methods. We augmented the data and constructed a matrix D (the document-term matrix) where each element $D_{ij}$ contains the weight of term $w_j$ in document $d_i$. Then, using topic modeling, we calculated the relevance of each topic $t_k$ for each document $d_i$ and the relevance of each term $w_j$ for each topic $t_k$. Finally, we applied a grid search and topic coherence to choose the best models and evaluate the external corpus results. In the topic filter and selection phase, we also processed the resulting topics by merging similar or removing topics with few documents. These results are presented in Section~\ref{sec:results}, while Fig.~\ref{fig:pipeline} illustrates the code-clustering pipeline. External evaluation is not depicted in this overview.

% \end{figure}

\subsubsection{Data Transformation}
\label{sec:bag_of_word}
\begin{figure}[h!]
    \centering
    \subfloat[Original code snippet]{\includegraphics[width=0.6\linewidth]{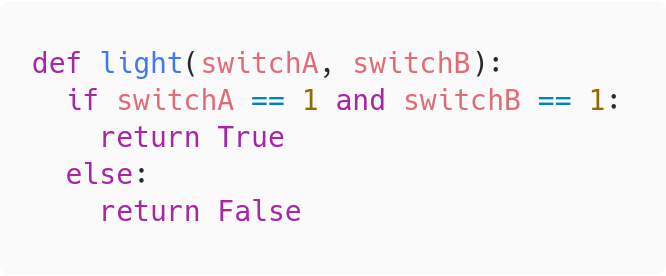}} \\
    \subfloat[Augmented code snippet]{\includegraphics[width=\linewidth]{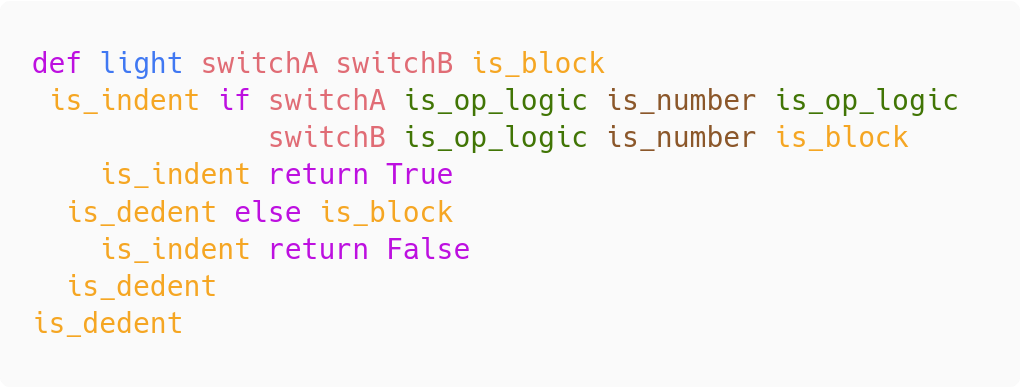}}
\caption{A code snippet and its augmented version. The augmented version will be processed in the augmented tokenizer, whereas the regular one will be processed in the standard tokenizer.}
\label{fig:code_snippets}
\end{figure}

In this application, the CS1 code solutions written in Python are considered documents. The document-term matrix creation process starts by splitting each code snippet into words. The first proposed tokenizer includes only split word tokens. Henceforth this tokenizer will be referred to as the \textit{standard tokenizer}. As stated in the related work section: 1) the LDA usually does not perform well on short texts and 2) augmenting the corpus by adding the text's structure on semi-supervised documents demonstrated improved results. We propose a new tokenizer to augment the standard tokenizer with extra features and refer to it as the \textit{augmented tokenizer}. The augmented tokenizer parses the code and makes special annotations by adding extra features if the token is a number, an array (or a list), a dictionary, a string, a logical (or arithmetic) operator, a class method, or indentation. The word itself is added to the document-term matrix if the token is a reserved word. Besides adding single tokens, this tokenizer also considers bigrams and trigrams. Although the document-term matrix does not consider the terms' order, this can be enforced by adding n-grams as a matrix feature. For example, the code snippet in Fig.~\ref{fig:code_snippets}(a) is first transformed to its augmented version (see Fig.~\ref{fig:code_snippets}(b)). Then, every single word, including the bigrams and the trigrams, are added as tokens to the document-term matrix. Table~\ref{tab:document_term_matrix} presents some examples of the document-term matrix terms, and, in total, the document is tokenized into 75 terms.

%%%%%%%%%%%%%%%%%%%% Figure/Image No: 1 starts here %%%%%%%%%%%%%%%%%%%%

%%%%%%%%%%%%%%%%%%%% Table No: 2 starts here %%%%%%%%%%%%%%%%%%%%

\begin{table}[h!]
 \centering
 \caption{Example of Document-Term Matrix. A Set of Terms from the Complete Document-Term Matrix after Augmenting and Tokenizing the Document from Fig.~\ref{fig:code_snippets}}
\begin{tabular}{|l|c|}
\hline 
%row no:1
Terms  & Count \\
\hline 
is\_block & 3 \\
\hline
is\_indent & 3 \\
\hline
is\_number & 2 \\
\hline
is\_op\_logic & 3 \\
\hline
if & 1 \\
\hline
is\_op\_logic is\_number & 2 \\
\hline
is\_block is\_indent & 3 \\
\hline
is\_op\_logic is\_number is\_op\_logic & 1 \\
\hline
\end{tabular}
\label{tab:document_term_matrix}
\end{table}

%%%%%%%%%%%%%%%%%%%% Table No: 2 ends here %%%%%%%%%%%%%%%%%%%%

After the document-term matrix creation, we applied some transformations to enhance document representation and decrease matrix sparsity. First, we removed tokens with document frequency below a fixed threshold to perform feature selection. This threshold was determined using a hyperparameter grid search ranging from 5\% to 50\% with a 5\% step. Second, we decided how to count a token frequency: either the token is counted once per document (binary appearance) or every time it appears. Finally, some tokens may be more important than others. For example, term frequency by inverse document frequency (TF-IDF)~\cite{Salton1986, Zhang2011} recalculates the tokens' weights by balancing two factors: 1) a term that occurs in many documents should not be as important as a more exclusive term, since it does not characterize documents well. Also, 2) a term that appears in a small number of documents may only be particular to those documents and not enough to distinguish a topic. Yan \textit{et al.}~\cite{Yan2012} propose another way of recalculating the terms' weights. Their method (called NCut) comes from the normalized cut problem on term affinity graphs. This weighting scheme modifies terms' counts based on terms cooccurrence and not on document frequency. Their experiments show NMF's performance increase on short text clustering using the NCut weighting scheme.

\subsubsection{Topic Extraction}
\label{sec:topic_extraction}
As stated earlier, after document processing, a document-term matrix $D$ is generated. The matrix rows represent points in an $\mathbb{R}^n$ feature space, where $n$ is the total number of terms, and each term $w_j$ corresponds to a dimension. It becomes a classical clustering problem where we expect similar documents to be in surrounding regions in space. So, clustering algorithms like K-means, hierarchical clustering, and nearest neighbors are applicable here. However, for topic modeling tasks, algorithms like the NMF~\cite{Lee1999, Cichocki2009} and the LDA~\cite{Blei.DavidM.2003} are effective since they interpret terms' counts as a set of visible variables generated from a set of hidden variables (topics)~\cite{OCallaghan2015,Chen2019}. Accordingly, the documents can be modeled as a distribution of topics and topics as a distribution of terms. We used two topic modeling techniques:

\begin{enumerate}
	\item NMF: a matrix factorization technique with a particular property of only allowing non-negative values in its entries, which is well-suited for human-interpretability~\cite{Lee1999}.
% 	In the topic modeling task, the topics are modeled as linear combinations of terms (a sum of each term weighted by its corresponding importance) and the documents as a weighted-sum of the topics it belongs to. 

	\item LDA: a generative probabilistic model that describes how to create documents in a collection. Once you have a dataset, a group of already written documents, we find the distributions that create these documents. The LDA algorithm tries to backtrack this probabilistic model to find a set of topics that are likely to have generated the dataset~\cite{Steyvers2010}. To generate a document, we sample from two distributions using the following iterative process: 
	\begin{enumerate}
	    \item We sample a topic for the given document (a document is a distribution of topics).
        \item We sample a  term from the topic sampled in step a) (a topic is a distribution of terms).
    \end{enumerate}
\end{enumerate}

\subsubsection{Topic Filter and External Evaluation}
\label{sec:topic_filter}
Given several document-term matrix creation options and two different topic modeling methods, we need to find the best set of hyperparameters. There are strategies in the literature to find a near-optimal set of models' hyperparameters, such as manual search, grid search, and random search~\cite{BergstraJAMESBERGSTRA2012}. Although random search demonstrates promising results in general machine learning tasks~\cite{BergstraJAMESBERGSTRA2012}, Chuang \textit{et al.}~\cite{chuang2013topic} and Wang and Blei~\cite{wang_collaborative_2011} results were competitive using grid search in topic modeling tasks. We chose to use a grid search approach. There was a prior manual stage to define the regions in which grid search would act. Since the dataset is not large and the number of hyperparameters to try is not extensive, it is efficient to run an exhaustive search combining hyperparameters. In total, there are 1680 possible combinations: 10 minimum document frequencies (ranging from 5\% to 50\% with 5\% step increment), 2 binary appearance options, 3 token weights (counts) transformation possibilities (none, TF-IDF, and NCut), 2 clustering methods (LDA and NMF), and 14 number of clusters (i.e., $10 \times 2 \times 3 \times 2 \times 14 = 1680$). The grid search was set to search between 2 and 15 clusters (the upper bound is based on the number of concepts from Table~\ref{tab:List of CS1 Concepts} in Section~\ref{sec:topics_context}).

To determine whether topics are well-defined, we can use topic coherence and pointwise mutual information (PMI) metrics, which correlated well with human-interpretability~\cite{mimno_optimizing_nodate, aletras_evaluating_nodate, lau_machine_2014}. As explained in Section~\ref{sec:topic_extraction} (topic extraction), when using NMF or LDA, each topic is mapped to a list of top-N words that best define the topic. Topic coherence calculates the ratio between the cooccurrence of these top-N words and their total occurrence. The assumption is that the words that best characterize a topic often appear together if a topic is well-defined. This paper applied two types of topic coherence metrics: UCI~\cite{newman_automatic_2010} and UMass~\cite{mimno_optimizing_nodate}. The UCI metric based on PMI is calculated using an external validation source. The PMI can be substituted using normalized PMI (NPMI) to better correlate with humans' ratings~\cite{aletras_evaluating_nodate}. The UMass metric uses the conditional probability of one word occurring given that one other high-ranked word occurred and can be measured using the modeled corpus, without depending on an external reference corpus. We used the UMass coherence to choose the best models since it is an internal validation metric (it only evaluates the clustered data). To assess the models, we used an external dataset with the UCI NPMI metric.

Defining $P(w_i)$ as the probability of the term $w_i$ occurring and $P(w_i, w_j)$ as the probability of terms $w_i$ and $w_j$ cooccurring, we calculated the coherence for a single topic $t_k$ using (\ref{eq:umass}), (\ref{eq:uci}), and (\ref{eq:npmi}). In this paper, the topic coherence for a single topic was calculated using top-5 and top-10 terms. After calculating each topic's coherence in a single hyperparameter combination, this combination's coherence was reported as the median of all topic coherence.

\begin{equation}
    C_{UMass}(W) = \sum_{i=1}^{N} \sum_{j=1}^{N} log \frac{P(w_i, w_j) + \epsilon}{P(w_i)}
\label{eq:umass}
\end{equation} 
\begin{equation}
    C_{UCI}(W) = \sum_{i=1}^{N} \sum_{j=1}^{N} NPMI(w_i, w_j)
\label{eq:uci}
\end{equation}
\begin{equation}
    NPMI (W) = \frac{log \frac{P(w_i, w_j) + \epsilon}{P(w_i) P(w_j)}}{-log(P(w_i, w_j) + \epsilon)}
\label{eq:npmi}
\end{equation}
where $ W = (w_1, w_2, ..., w_N)$ are the top-N terms for calculating the coherence. An $\epsilon$ value of 0.01 was used to avoid taking a zero logarithm.

We performed hard-assignment to cluster documents by topic by assigning each document to the topic with the most relevance (weight) in the document-topic matrix. The hard-assignment was achieved with minimal loss of information when a topic strongly characterized a document. In addition to assigning documents to topic clusters, the set of features/terms that best characterize each cluster/topic were extracted for further analysis. 

As a final step, we also merged semantically similar topics and removed those not containing relevant information.

\subsection{Topics Contextualization}
\label{sec:topics_context}
To relate concepts and topics, we first defined the most commonly seen concepts in CS1 exercises. The following four references were used to create a list of concepts commonly used in CS1 courses:

\begin{enumerate}
	\item Computer Science Curricula 2013~\cite{ACM2013}: a document jointly built by the Association of Computer Machinery and the IEEE Computer Society. The document recommends curricular guidelines for computer science education, which we used as the main concept list. We used the papers in items~\ref{b}, \ref{c}, and~\ref{d} to improve it.

	\item Exploring programming assessment instruments: A classification scheme for examination questions~\cite{Sheard2011}: creates a classification scheme characterizing exam questions by their concept areas, question style, and skills a student needs to solve them. We used the list of the proposed concepts as a second source to enhance the main list.
	\label{b}

	\item Reviewing CS1 exam question content~\cite{Petersen2011}: this paper used nine experienced CS1 instructors to review the concepts required in examinations from different North American institutions. It created a list of concepts using the intersection of three other experiments~\cite{Schulte2006, Robins2006, Goldman2010}.
	\label{c}

	\item Identifying challenging CS1 concepts in a large problem dataset~\cite{Cherenkova2014}: this paper identifies the most challenging CS1 concepts for students based on 266,852 web-based code-writing student responses. Their concept list is based on the course structure from the CS1 class where they experimented.
	\label{d}
\end{enumerate}

Table~\ref{tab:List of CS1 Concepts} shows the final list of consolidated concepts.

%%%%%%%%%%%%%%%%%%%% Table No: 1 starts here %%%%%%%%%%%%%%%%%%%%
\begin{table}[h!]
 \centering
 \caption{List of CS1 Concepts}
\begin{tabular}{|l|l|l|}
\hline 
% \hline
%row no:1
1. Syntax &  6. Logic &  11. Conditional \\
\hline
%row no:2
2. Assignment & 7. Data type: string & 12. Loop \\
\hline
%row no:3
3. Data type: number & 8. Data type: array & 13. Nested loop \\
\hline
%row no:4
4. Data type: boolean & 9. Data type: tuple & 14. Function \\
\hline
%row no:5
5. Math & 10. Data type: dict & 15. Recursion \\
\hline
\end{tabular}
\label{tab:List of CS1 Concepts}
\end{table}

Then, to interpret the meaning of the topics, we asked 14 professors to perform three tasks. The professors (with 2 to 20 years of teaching experience) teach CS1 or other programming-related subjects.
\begin{itemize}
    \item \textbf{Theme identification:} we present some code snippets belonging to the topic and found essential tokens for each topic. The professors were asked to label each topic with free-text descriptions. We tokenized the descriptions and counted the terms. We also created the topic titles based on the terms that appeared more frequently in the descriptions.
    \item \textbf{Concept identification:} each professor was asked to associate up to three concepts (from the 15 available in Table~\ref{tab:List of CS1 Concepts}) to 15 randomly assigned code snippets. Then, we aggregated the concepts related to each code to the associated topics. In this way, it is possible to understand the most relevant concepts in each topic and calculate how well the concepts inside a cluster agree to them. A similar approach was used by Lan \textit{et al}.~\cite{lan2014} to contextualize concept clusters.
    \item \textbf{Intruder identification:} Chang \textit{et al.}~\cite{chang_reading_2009} proposed quantitative methods to analyze the interpretability of the latent space created using topic modeling techniques. Their paper proposed a method to identify an intruder topic given a document. We adapted this method because we hard-assigned each document to a single topic. For our analysis, the professors were asked to identify the intruder document given a topic. This approach has been used by Mahmoud and Bradshaw~\cite{mahmoud2017semantic} to assess the quality of their topic modeling approach on Java-based systems.
\end{itemize}
% \chapter{Results and Discussion}
\section{Results and Discussion}
\label{sec:results}
We run each hyperparameter combination from the 1680 possibilities 10 times and calculated their average coherence and standard deviation. Next, the two best-ranked results are analyzed. They were calculated using Fagin’s algorithm~\cite{fagin1996combining} for top-5 and top-10 terms UMass coherence. Table~\ref{tab:exp1_params} shows the set of hyperparameters for each experiment. 

%%%%%%%%%%%%%%%%%%%% Table No: 3 starts here %%%%%%%%%%%%%%%%%%%%
\begin{table}[h!]
\centering
\caption{Set of HyperParameters for Each Experiment}
\begin{tabular}{l|c|c|c|c|c}
\hline
%row no:1
 & Min DF & Binary & Vectorizer & Method & \# of clusters \\
\hline
%row no:2
Exp. 1 & 0.35 & True & NCut & NMF & 7 \\
Exp. 2 & 0.05 & True & Count & LDA & 12 \\
\hline
\end{tabular}
\label{tab:exp1_params}
\end{table}
%%%%%%%%%%%%%%%%%%%% Table No: 3 ends here %%%%%%%%%%%%%%%%%%%%

\subsection{Experiment 1}
\begin{figure}[h!]
    \centering
    \includegraphics[width=0.4\textwidth]{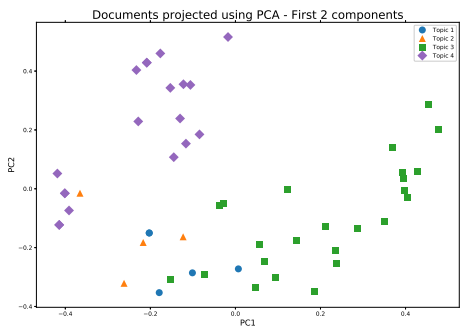}
	\caption{Documents projected into the first two dimensions using principal component analysis. Points with the same color and marker belong to the same cluster.
}
	\label{fig:exp1_pca}
\end{figure}

After the document-term matrix factorization, we hard-assigned each document to the topic with the highest relevance (highest weight in the document-topic distribution). Table~\ref{tab:exp1_topic_distribution} shows the number of documents assigned per topic. After assigning each document to its related topic in this experiment, the documents are only assigned to four of the seven topics. Fig.~\ref{fig:exp1_pca} shows the documents projected to two dimensions using principal component analysis (PCA).

%%%%%%%%%%%%%%%%%%%% Table No: 4 starts here %%%%%%%%%%%%%%%%%%%%
\begin{table}[h!]
\centering
\caption{Number of Documents per Topic in Experiment 1}
\begin{tabular}{c|c|c|c|c|c|c|c}
\hline
%row no:1
Topics & 1 & 2 & 3 & 4 & 5 & 6 & 7\\
\hline
%row no:2
\# of documents & \textbf{5} & \textbf{4} & \textbf{26} & \textbf{19} & 0 & 0 & 0\\
\hline
\end{tabular}
\label{tab:exp1_topic_distribution}
\end{table}
%%%%%%%%%%%%%%%%%%%% Table No: 4 ends here %%%%%%%%%%%%%%%%%%%%

%%%%%%%%%%%%%%%%%%%% Figure/Image No: 4 starts here %%%%%%%%%%%%%%%%%%%%

%%%%%%%%%%%%%%%%%%%% Figure/Image No: 4 Ends here %%%%%%%%%%%%%%%%%%%%

\begin{figure}[h!]
    \includegraphics[width=0.4\textwidth]{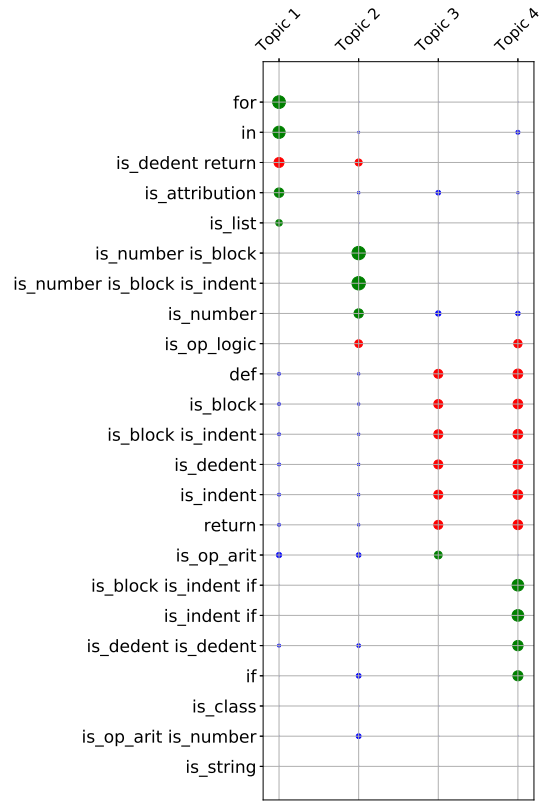}
	\caption{Term importance for the four most populated topics. Each row represents a term, the size of the point corresponds to the term's weight for the topic, and red points are the points above the 75th percentile of all weights. The green points denote words above the 75th percentile limit on only one of the four topics.}
	\label{fig:exp1_term_per_topic}
\end{figure}

Using a minimum document frequency of 35\%  kept only 23 valid terms. Fig.~\ref{fig:exp1_term_per_topic} shows the essential terms per topic where the terms that are exclusively important for a single topic (a term is vital if it is above the 75th percentile of all weights) are denoted in green. In this plot, topics 3 and 4 share almost all terms. By adjusting the document-term matrix values using the NCut vectorizer, the factorization split topics 3 and 4 using the conditional \textit{if} term. Topic 4 is exclusive for code snippets that are solved using conditional statements, whereas topic 3 comprises the opposite. 

%%%%%%%%%%%%%%%%%%%% Figure/Image No: 5 starts here %%%%%%%%%%%%%%%%%%%%
%%%%%%%%%%%%%%%%%%%% Figure/Image No: 5 Ends here %%%%%%%%%%%%%%%%%%%%

\begin{figure}[h!]
    \centering
    \includegraphics[width=0.5\textwidth]{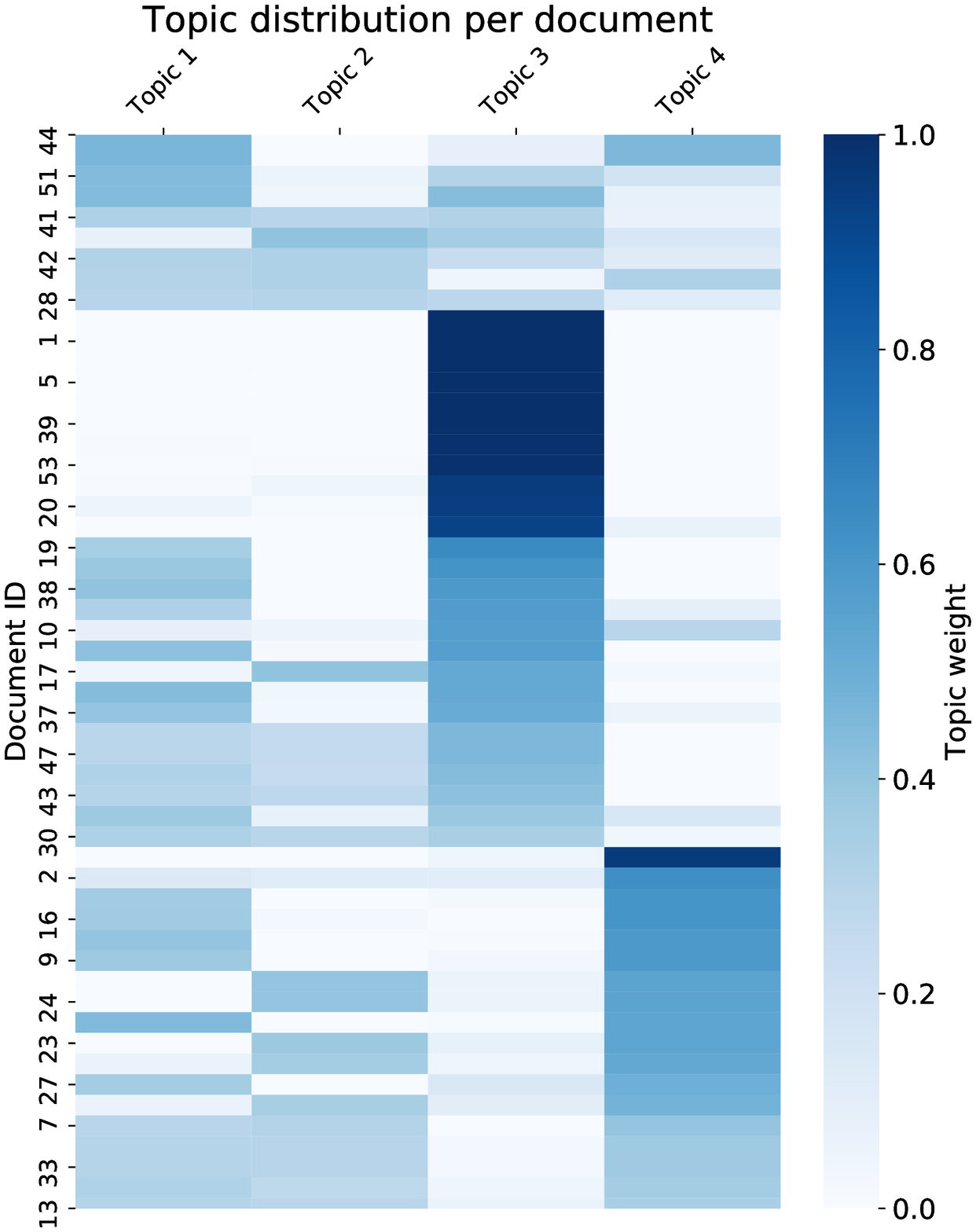}
    \caption{Topic distribution per document. Darker cells indicate a better description of the document by the topic.}
    \label{fig:exp1_document_topic}
\end{figure}

Fig.~\ref{fig:exp1_document_topic} shows the topic distribution per document. As explained in Section~\ref{sec:topic_extraction} (topic extraction), distribution over topics describes a document. Darker cells imply that the topic characterizes a document better. As stated before, if a topic strongly characterizes a document, then we can hard-assign it to a single topic. However, Fig.~\ref{fig:exp1_document_topic} shows that most documents assigned to topics 1 and 2 (top part of the plot) spread throughout the topics. It suggests we have to combine the most important terms for each topic to interpret these code snippets.

%%%%%%%%%%%%%%%%%%%% Figure/Image No: 6 starts here %%%%%%%%%%%%%%%%%%%%
%%%%%%%%%%%%%%%%%%%% Figure/Image No: 6 Ends here %%%%%%%%%%%%%%%%%%%%

Analyzing the code snippets from topic 1, they combine for-loops with conditional statements. Topic 2 is a mixture containing the code snippets that do not belong to any other topic.

\begin{figure}[ht!]
    \centering
	\includegraphics[width=0.4\textwidth]{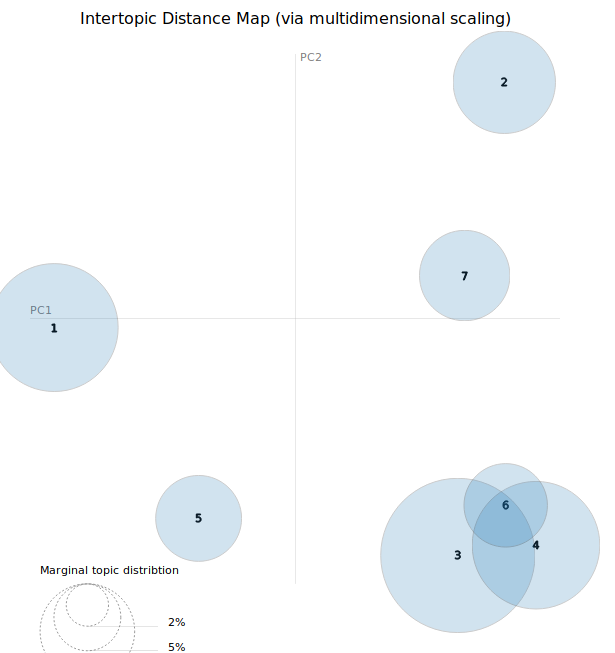}
	\caption{Topics are represented as circles proportional to the number of terms whose weights are most associated with the topic and the distance between the circles is the intertopic distance.}
	\label{fig:exp1_vistool}
\end{figure}

Fig.~\ref{fig:exp1_vistool}, using the LDAVis tool~\cite{Sievert2014}, calculates the topics' distance and projects them to 2D using principal coordinate analysis. Topics 3 and 4 are located close to each other, and they correspond to 45\%  of the terms and 83\%  of the documents. Fig.~\ref{fig:exp1_vistool} also validates that these topics are not that different when their crucial terms are analyzed. Still, the conditional statements that characterize topic 4 are enough to produce a linearly separable 2D data projection, except for a few outliers, as shown in Fig.~\ref{fig:exp1_pca}.

%%%%%%%%%%%%%%%%%%%% Figure/Image No: 7 starts here %%%%%%%%%%%%%%%%%%%%

\FloatBarrier

%%%%%%%%%%%%%%%%%%%% Figure/Image No: 7 Ends here %%%%%%%%%%%%%%%%%%%%
% \clearpage
\subsection{Experiment 2}
\label{ssec:exp_2}
%  Table~\ref{tab:exp2_params} describes the set of hyper-parameters for this experiment.
 
% %%%%%%%%%%%%%%%%%%%% Table No: 5 starts here %%%%%%%%%%%%%%%%%%%%
% \begin{table}[h!]
% \centering
% \begin{tabular}{c|c|c|c|c}
% \hline
% %row no:1
% Min DF & Binary & Vectorizer & Method & \# of clusters \\
% \hline
% %row no:2
% 0.05 & True & Count & LDA & 12 \\
% \hline
% \end{tabular}
% \caption{Set of hyper-parameters for experiment 2}
% \label{tab:exp2_params}
% \end{table}

%%%%%%%%%%%%%%%%%%%% Table No: 5 ends here %%%%%%%%%%%%%%%%%%%%

\begin{figure}[h!]
\centering
    \includegraphics[width=0.5\textwidth]{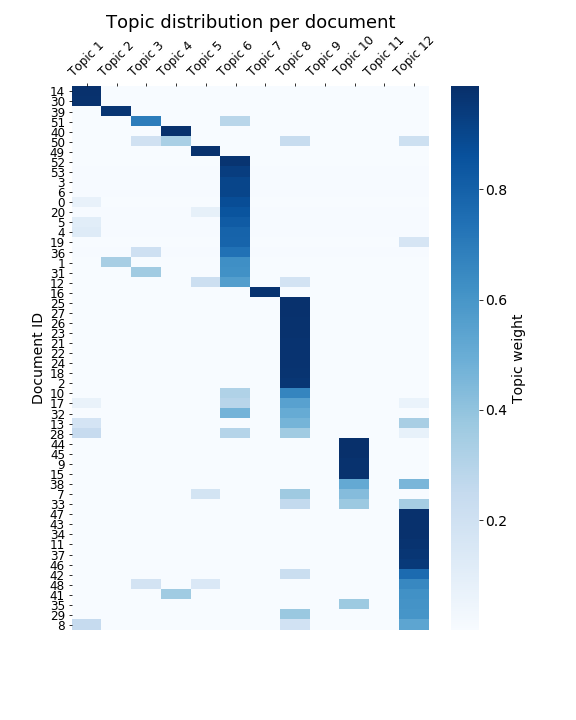}
	\caption{Topic distribution per document. Darker cells indicate a better description of the document by the topic.}
		\label{fig:exp2_document_topic}
\end{figure}

 Table~\ref{tab:exp2_topic_distribution} shows the number of assigned documents per topic with hyperparameters combination producing a more uniform grouping scheme than the previous one. Although we initially set 12 clusters, two of them (topics 9 and 11) are empty after assigning each document to the topic with the highest relevance (weight). Topics 6, 8, 10, and 12 have the largest number of documents. Fig.~\ref{fig:exp2_document_topic} shows the topic per document distribution where the topics better characterize each document.
 
 %%%%%%%%%%%%%%%%%%%% Table No: 6 starts here %%%%%%%%%%%%%%%%%%%%
 \begin{table}[h!]
\centering
\caption{Number of Documents per Topic in Experiment 2}
\resizebox{\columnwidth}{!}{%
\begin{tabular}{c|c|c|c|c|c|c|c|c|c|c|c|c}
\hline
%row no:1
Topics & 1 & 2 & 3 & 4 & 5 & 6 & 7 & 8 & 9 & 10 & 11 & 12\\
\hline
%row no:2
\makecell{\# of \\ documents} & 2 & 1 & 1 & 2 & 1 & \textbf{13} & 1 & \textbf{14} & 0 & \textbf{7} & 0 & \textbf{12}\\
\hline
\end{tabular}}
\label{tab:exp2_topic_distribution}
\end{table}
%%%%%%%%%%%%%%%%%%%% Table No: 6 ends here %%%%%%%%%%%%%%%%%%%%

%%%%%%%%%%%%%%%%%%%% Figure/Image No: 8 starts here %%%%%%%%%%%%%%%%%%%%

%%%%%%%%%%%%%%%%%%%% Figure/Image No: 8 Ends here %%%%%%%%%%%%%%%%%%%%

The complete term per topic plot for this experiment is omitted because it is long and challenging to read. Using a minimum document frequency of 5\% increased the number of terms to 236.

%%%%%%%%%%%%%%%%%%%% Figure/Image No: 9 Ends here %%%%%%%%%%%%%%%%%%%%

\begin{figure}[h!]
\centering
    \includegraphics[width=0.41\textwidth]{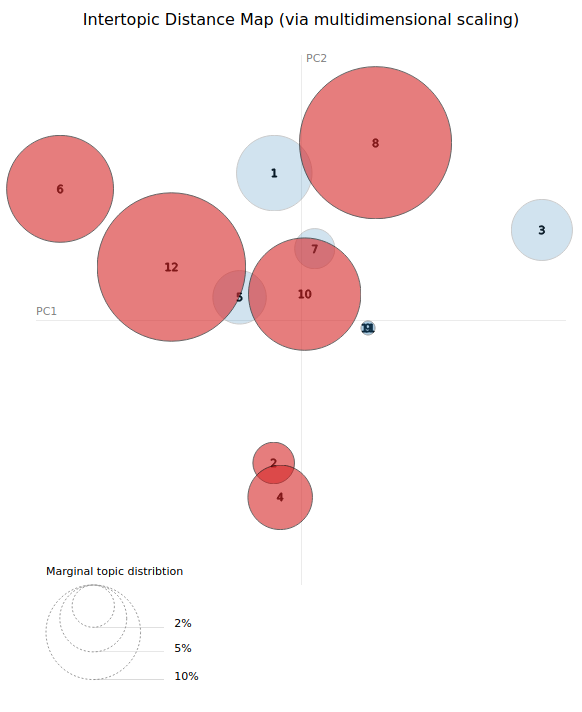}
	\caption{Intertopic distance map for experiment 2. Topics are represented as circles proportional to the number of terms whose weights are most associated with the topic and the distance between the circles is the intertopic distance. Topics in red are further discussed in detail.}
	\label{fig:exp2_vistool}
\end{figure}

%The main topics 6, 8, 10, and 12 correspond to 85$\%$  of the documents and 77.4$\%$  of the terms. These topics do not overlap as much as the topics from Experiment 1, as shown in 
Fig.~\ref{fig:exp2_vistool} shows the intertopic map. The main topics 6, 8, 10, and 12 correspond to 85$\%$  of the documents and 77.4$\%$  of the terms and we do not observe any main topic overlap in this plot. The next subsections analyze these main topics in detail. Topics 2 and 4 will also be analyzed since they occupy a different space on the map. This step belongs to the topic filter and selection phase from the code-clustering pipeline depicted in Fig.~\ref{fig:pipeline}. After hard-assigning the documents to the clusters (removing topics 9 and 11), merging topics 2 and 4, and removing topics with a few documents (less than three documents per topic: topics 1, 3, 5, and 7), it resulted in five conceptual clusters (six from the original topics in total) to be analyzed in detail.

%%%%%%%%%%%%%%%%%%%% Figure/Image No: 10 starts here %%%%%%%%%%%%%%%%%%%%
%%%%%%%%%%%%%%%%%%%% Figure/Image No: 10 Ends here %%%%%%%%%%%%%%%%%%%%

\begin{figure}[h!]
    \subfloat[Topic 2 ]{\includegraphics[width=0.5\linewidth]{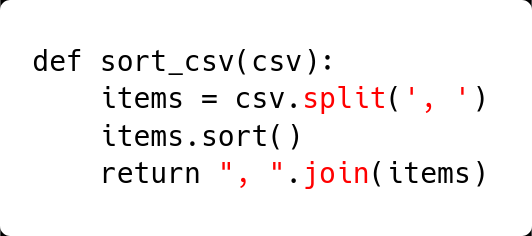}} \\
    \subfloat[Topic 4]{\includegraphics[width=0.6\linewidth]{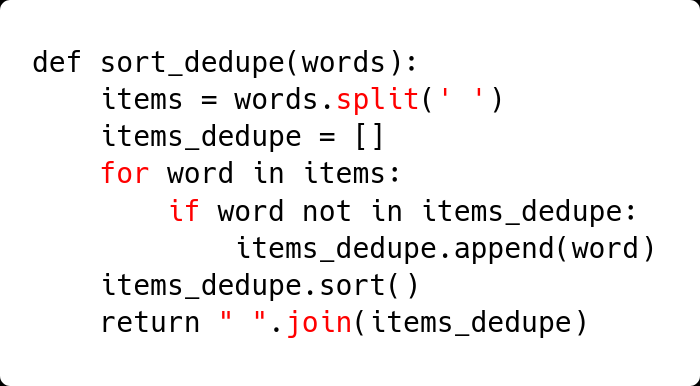}} \\
    \subfloat[Topic 6]{\includegraphics[width=0.65\linewidth]{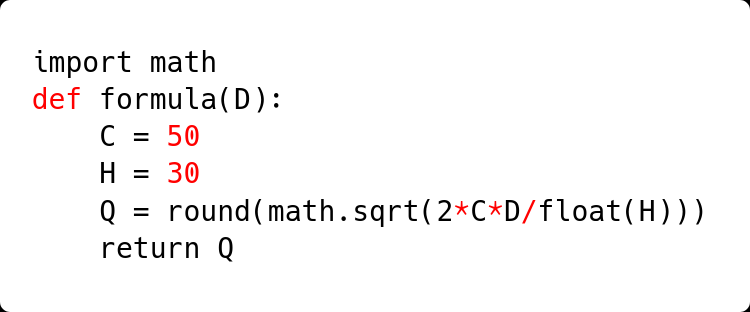}} \\
    \subfloat[Topic 8]{\includegraphics[width=0.9\linewidth]{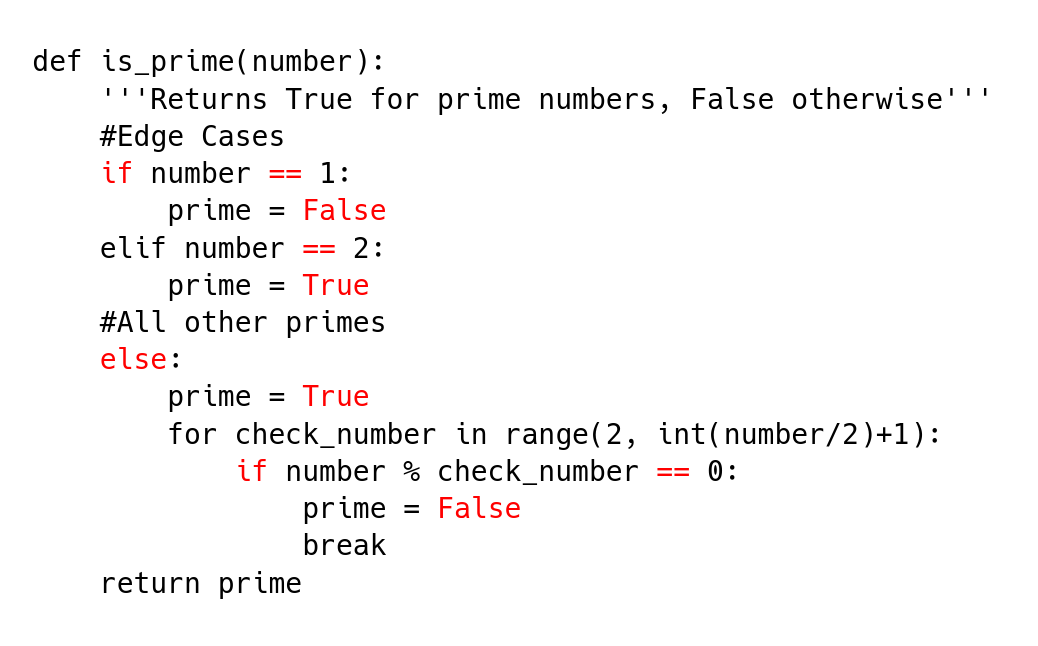}} \\
    \subfloat[Topic 10]{\includegraphics[width=0.6\linewidth]{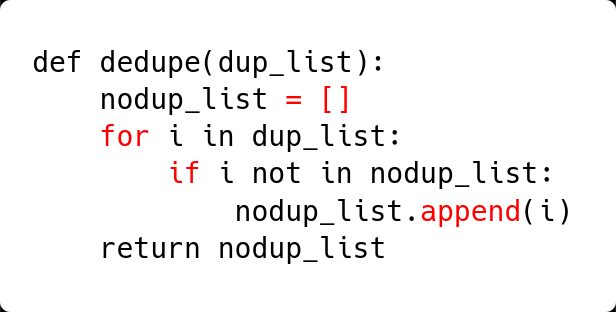}} \\
    \subfloat[Topic 12]{\includegraphics[width=0.6\linewidth]{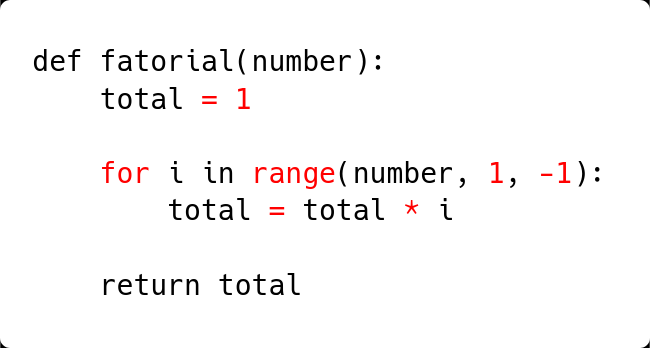}} \\
\caption{Example of code snippets belonging to each topic. In red are the relevant terms for each topic.}
\label{fig:code_snippet_topic}
\end{figure}

Using the Sievert and Sheley relevance metric implemented in the LDAVis tool~\cite{Sievert2014}, the top-30 most relevant terms per topic were extracted. Table~\ref{tab:exp2_topwords} shows the five most relevant terms.
% after finding and merging similar unigrams, bigrams and 3-grams structures.
We wrote a description for each topic after analyzing the essential terms and code snippets from each class. Fig.~\ref{fig:code_snippet_topic} shows some examples of code snippets for each topic, highlighting the terms used to define them.

\begin{itemize}
	\item Topic 8 is strongly characterized by conditional statements, logical operators, and Boolean values.

    \item Topic 6 does not seem to have a clear definition by just inspecting its terms. Indentation terms appear among the most relevant ones. By analyzing the code snippets, topic 6 comprises code with one indentation structure (simple coding structures), sometimes without assigning variables to solve the exercise.
    
    \item Topic 12 is characterized by for-loops, especially range loops combined with arithmetic operations.

    \item Topics 2 and 4 present for-loops, conditionals, and lists, and these tools are used to perform string operations. A strong mark is the presence of the auxiliary functions split and join.
    
    \item Topic 10 is about lists and their usual operations: for-loops, conditionals, and appending elements.

\end{itemize}

\begin{table}[h!]
\centering
\caption{Five Most Relevant Terms for Each of the Analyzed Topics. The Terms Starting with `Is' Are the Special Annotated Terms Explained in Section~\ref{sec:bag_of_word} (Data Transformation)}
\begin{tabular}{|l|l|l|l|l|}
\hline
\multicolumn{1}{|l|}{\begin{tabular}[c]{@{}l@{}}\textbf{Topics} \\ \textbf{2 \& 4}\end{tabular}} & \textbf{Topic 6} & \textbf{Topic 8}      & \textbf{Topic 10}        & \textbf{Topic 12}        \\ \hline
split                  & is\_op\_arit     & is\_op\_logic & append          & range           \\ \hline
is\_string             & is\_indent       & if            & is\_list        & is\_op\_arit    \\ \hline
join                   & def              & else          & for             & for             \\ \hline
for                    & is\_number       & True          & is\_attribution & is\_number      \\ \hline
if                     & len              & False         & if              & is\_attribution \\ \hline
\end{tabular}
\label{tab:exp2_topwords}
\end{table}

% \FloatBarrier

\begin{figure}[h!]
\centering
    \includegraphics[width=0.45\textwidth]{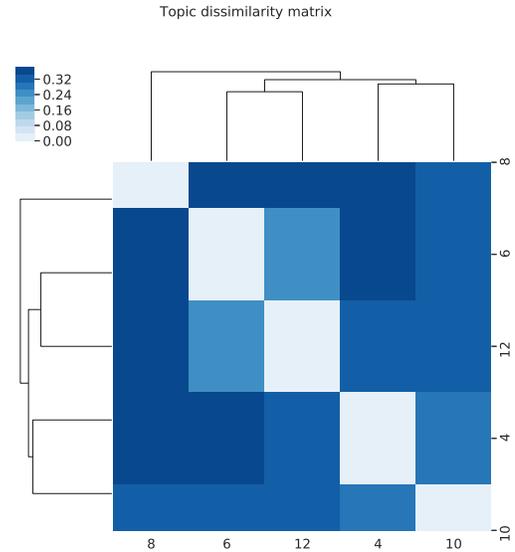}
	\caption{Topic dissimilarity matrix. The darker the cell more distant the topics are from each other.}
	\label{fig:topic_distances}
\end{figure}

 To better understand the topics' differences, we performed a topic dissimilarity analysis shown in Fig.~\ref{fig:topic_distances}, representing the real distances of the topics, not reduced to two dimensions as in Fig.~\ref{fig:exp2_vistool}. In this analysis, topic 8 (conditional) is the furthest from all others. Topic 10 is central, being slightly closer to topic 4 than to topic 12. This result is reasonable as it contains elements between these two topics (numeric and range loops vs. loops with strings). Table~\ref{tab:exp2_topwords} shows their differences as being the data types, which can be observed in the most important terms (``range'' and ``is\_number'' for topic 12, ``append'' and ``is\_list'' for topic 10 and ``split'', and ``is\_string'' for topics 2 and 4).

%%%%%%%%%%%%%%%%%%%% Table No: 7 starts here %%%%%%%%%%%%%%%%%%%%

%%%%%%%%%%%%%%%%%%%% Table No: 7 ends here %%%%%%%%%%%%%%%%%%%%

%%%%%%%%%%%%%%%%%%%% Figure/Image No: 11 starts here %%%%%%%%%%%%%%%%%%%%

We also analyzed how the test dataset fits into the six (including topics 2 and 4) valid topics to understand if the topics are representative of the possible concepts present in an unseen code. We assigned each code to the topic with the highest weight as we did for the training set. Table~\ref{tab:exp2_test_topic_distribution} shows the number of assigned documents per topic. Except for two documents, all the others belong to one of the six valid topics. It confirms that the different topics (the ones considered invalid) detect specific code traits and not their general concepts. It is important to notice that topic modeling is a soft clustering technique: a document has a probability of belonging to each topic and can be associated with more than one. So, a document can be related to the main topic with its specificity related to minor ones.

 \begin{table}[h!]
\centering
\caption{Number of Test Set Documents per Topic for Experiment 2}
\resizebox{\columnwidth}{!}{%
\begin{tabular}{c|c|c|c|c|c|c|c|c|c|c|c|c}
\hline
%row no:1
Topics & 1 & 2 & 3 & 4 & 5 & 6 & 7 & 8 & 9 & 10 & 11 & 12\\
\hline
%row no:2
\makecell{\# of \\ documents} & 2 & \textbf{4} & 0 & \textbf{1} & 0 & \textbf{23} & 0 & \textbf{17} & 0 & \textbf{3} & 0 & \textbf{15}\\
\hline
\end{tabular}}
\label{tab:exp2_test_topic_distribution}
\end{table}

%%%%%%%%%%%%%%%%%%%% Figure/Image No: 15 Ends here %%%%%%%%%%%%%%%%%%%%

\subsection{Coherence Evaluation}
Both experiments were analyzed using the UCI coherence metric with NPMI~\cite{aletras_evaluating_nodate}, as described in Section~\ref{sec:topic_filter} (topic filter and external evaluation), to validate how well the proposed methodology performs in an external dataset. Although AST trees have been used to cluster code, they do not provide an intuitive way to analyze the important features besides reading it. We compare our results with a K-means clustering method using the proposed augmented tokenizer and logistic regression to extract the important features per cluster as a baseline. We also compared our best results using the standard tokenizer instead of our proposed tokenizer. We used $k = 5$ for K-means since there were five main conceptual clusters found in the LDA. We ran each method 100 times and averaged their UCI coherence metric. Statistical difference was measured using the Mann--Whitney U test~\cite{mann_test_1947}, and all the results were statistically significant with $p < 0.001$. Table~\ref{tab:cuci_npmi} reports the mean and standard deviation for each experiment. In the UCI coherence with NPMI metric, the values are bounded between 1 and -1, where 1 means that the top words only occur together, zero means that they are distributed as expected under independence, and -1 means that they only occur separately. The UCI coherence for the standard tokenizer using the top-10 terms could not be measured because there were no important top-10 term pairwise combinations in this setting that appeared in at least one document. The NMF experiment with the augmented tokenizer considering the top-10 terms demonstrated the best UCI occurrence metric, followed by the LDA experiment, which had the best performance considering the top-5 terms.

\begin{table}[h!]
\centering
\caption{The Mean and Standard Deviation of UCI Coherence Using NPMI}
\begin{tabular}{l|l|l|}
\cline{2-3} & \multicolumn{2}{c|}{$C_{UCI}$ (NPMI)} \\ 
\cline{2-3} & Top-5 terms          & Top-10 terms         \\ \hline
\multicolumn{1}{|l|}{\begin{tabular}[c]{@{}l@{}}NMF (Experiment 1 with \\ augmented tokenizer)\end{tabular}} & 0.73 (0.03)          & \textbf{0.83 (0.02)} \\ \hline
\multicolumn{1}{|l|}{\begin{tabular}[c]{@{}l@{}}LDA (Experiment 2 with \\ augmented tokenizer)\end{tabular}} & \textbf{0.75 (0.06)} & 0.76 (0.04)          \\ \hline
\multicolumn{1}{|l|}{\begin{tabular}[c]{@{}l@{}}K-Means (with \\ augmented tokenizer)\end{tabular}}          & 0.55 (0.09)          & 0.53 (0.03)          \\ \hline
\multicolumn{1}{|l|}{\begin{tabular}[c]{@{}l@{}}LDA (best result for \\ standard tokenizer)\end{tabular}}    & 0.53 (0.03)          & ---                  \\ \hline
\end{tabular}
\label{tab:cuci_npmi}
\end{table}

\subsection{Discussion about Experiments 1 and 2}
Experiments 1 and 2 are the two best-ranked results on the top-5 and top-10 UMass coherence metric. Both experiments have different hyperparameters. Experiment 1 uses a minimum document frequency of 35\%, NCut to weight the terms, and then perform NMF to extract the topics. Experiment 2 uses a minimum document frequency of 5\%, regular count of words, and LDA to extract the topics. 

We found both experiments to have their main concepts in a few clusters (two main clusters in Experiment 1 and six main clusters in Experiment 2). The remaining clusters are associated with code specificity. In the case of Experiment 1, using NCut and a high document frequency threshold, the topic modeling from Experiment 1 focused on finding structures with high volume and cooccurrence rates, resulting in separation of the \textit{if/else} structure from the rest. The conditional structure was first separated from a hierarchical perspective, and the remaining structures were all grouped in a cluster. In Experiment 2, the conditional topic (topic 8) is also the furthest from the other topics. As shown in the hierarchical clustering of Fig. 9, this topic is the last one to be aggregated (or the first one to be separated). The common code snippets between the conditional clusters in each experiment also validate this result. From the 14 code snippets associated with the conditional topic (topic 8) in Experiment 2, 11 of them (79\%) belong to the conditional topic in Experiment 1 (topic 4). Therefore, Experiment 2 demonstrates more granularity than Experiment 1.

\subsection{Topics Contextualization}
Experiment 2 shows a better distribution of exercises and less overlap among topics than Experiment 1; we followed up the analysis with this experiment, which was the second-best result considering the $C_{UCI}$ with NPMI metric.
\begin{itemize}
    \item \textbf{Theme identification:} after tokenizing the labels given by the professors, we separated the most frequent tokens and manually created the topic titles, as follows. All the tokens in the presented titles appeared in at least 50\% of the labels.
        \begin{enumerate}
            \item Topics 2 \& 4: string manipulation,
            \item Topic 6: math functions,
            \item Topic 8: conditional structure,
            \item Topic 10: list loops,
            \item Topic 12: math loops.
        \end{enumerate}
    \item \textbf{Concept identification:} each professor was asked to associate up to three concepts (from the 15 available in Table~\ref{tab:List of CS1 Concepts}) to each presented code. Four professors analyzed each code. In 37 of the 54 code snippets, there was at least one concept in common between all four professors. In 53 of the 54 code snippets, at least one concept was common between three out of the four professors (75\%). Therefore, we decided to use the 75\% threshold of the agreement to relate the exercises' concepts. The concepts in each topic were aggregated to provide an overview of the main concepts needed to solve the cluster's problems, as summarized in Table~\ref{tab:topic_concept}. These results validate the topic themes defined in the previous task: the professors also elected each topic's main concept as a word in free-text labels. Notice that we performed the two tasks independently.

    % \begin{figure}[h!]
    %     \centering
    %     \includegraphics[width=0.5\textwidth]{media/evaluation.eps}
    %     \caption{ Relation between the found topics and the CS1 concepts. The top-3 concepts for each topic are highlighted in orange.}
    %   \label{fig:evaluation}
    % \end{figure}
    
\begin{table}[h!]
 \centering
 \caption{The relation between the Found Topics and the Main CS1 Concept Related to the Topic}
\begin{tabular}{|l|l|l|}
\hline
Topic                      & Main Concept         & \multicolumn{1}{|l|}{\begin{tabular}[c]{@{}l@{}}Agreement \\inside cluster\end{tabular}} \\ \hline
2 \& 4 - String manipulation    & 7. Data type: string & 66.7\%                   \\ \hline
6 - Math functions          & 14. Function         & 50\%                     \\ \hline
8 - Conditional structure  & 11. Conditional      & 86.7\%                   \\ \hline
10 - List loops            & 12. Loop             & 71.4\%                   \\ \hline
12 - Math loops            & 12. Loop             & 75\%                     \\ \hline
\end{tabular}
\label{tab:topic_concept}
\end{table}

    \item \textbf{Intruder identification:} four code snippets were presented in each of the five groups, three belonging to the same topic (randomly chosen from the topic pool) and an intruder (also randomly chosen from another topic). Fig.~\ref{fig:confusion_matrix} shows a confusion matrix, presenting how well the professors could distinguish the intruder in each topic. In this figure, each row sums to 1 and represents how often the professors correctly guess the intruder (the diagonal values) and how often the professors confuse the intruder (the intruder cluster is depicted in the columns). We can draw some insights from this analysis: 
    \begin{itemize}
        \item The ``conditional structure'' topic performed well, with the intruder code being identified 79\% of the time, meaning that identifying a code snippet from a different cluster can be done 4 out of 5 times. 
        \item The intruder code inside the ``math loops'' topic was identified 2 out of 3 times (64\%), being confused with ``list loops'' the last third of the time. These topics work on the same main concept, as seen in the concept identification task. 
        \item The same behavior is not seen for the ``list loops'' topic. This topic and the ``string manipulation'' topic present very similar behavior. They can be distinguished frequently (2x better than the 25\% random baseline), but we do not see a confusion pattern. 
        \item On the other hand, the ``math functions'' topic seemed to confuse the professors, being switched half of the time with the ``math loops'' topic. By backtracking the previous two tasks and analyzing them with this point of view, this topic's main concept was function, present in all exercises. As stated in Section~\ref{ssec:exp_2}, this topic does not have well-defined terms; indentation terms appeared among the most relevant. Although the indentations when programming in Python can indicate the difficulty of an exercise, they are not a natural human way of splitting them. Therefore, it is hard for humans to interpret this kind of clustering scheme.
    \end{itemize}

    \begin{figure}[h!]
        \centering
        \includegraphics[width=0.5\textwidth]{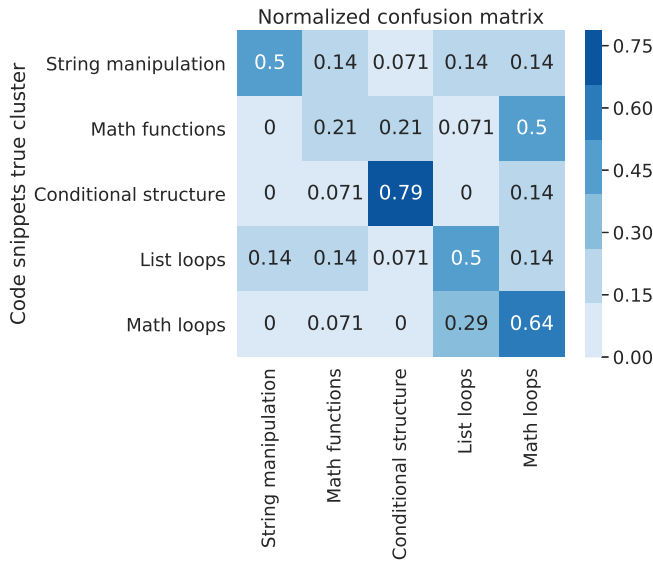}
        \caption{Normalized confusion matrix for the intruder-identification task.}
       \label{fig:confusion_matrix}
    \end{figure}
\end{itemize}
\section{Conclusions and Future Work}
\label{sec:conclusion}
Based on the evaluation metric, our proposed method found semantically related code-clustering schemes suitable for human-interpretability with minimal supervision, giving support for RQ1. The method is expected to provide semantics for large amounts of unannotated code. 

Although code clustering in the CS1 context has been widely applied using the AST trees, the advantages of working with topic modeling are the terms per topic results that may help experts better assess each cluster's contents. The methodology has also been shown to overcome the small-sized code snippets challenge by extending the tokenizer to augment the corpus with the code structure. The standard tokenizer could not create semantically related topics, but adding structural information, as features: indents and data types, and enforcing the order using n-grams enriched the code representation and found topics suitable for human-interpretability. For example, augmenting the corpus with structural information as indents/blocks (in Python, indents indicate how deep a block of code is; other languages like C++ and Java could count the number of ``\{'' and ``\}'') helps to separate single loops from nested loops. Combining trigrams (to enforce order) with structural information can distinguish subtle differences in precondition and postcondition loops. Notice that postcondition loops do not exist in Python, so we could not verify this specific assumption. In our dataset, we expect trigram tokens to be enough to capture these varieties because a typical CS1 solution does not have more than three or four nested structures. Still, it may limit our model in identifying large nested structures on more complex code. Also, even though there is a recursion concept in the concepts list, there was no exercise using this technique in our dataset to verify how it would be clustered.

In the second experiment, we set the best number of topics to 12, but only six (considering the merge of topics 2 and 4) were valid. Standard topic modeling techniques with the augmented tokenizer yielded the best results. The LDA-based clustering demonstrates better interpretable results, as shown by our detailed topic analysis. When tested with an unseen dataset, the six topics comprised the main concepts present in the test set. Except for a few exceptions, most of them relate to the six main topics, and a few of them account for code specificity.

To understand how humans could read and interpret the extracted relationships (RQ2), we asked 14 professors to contextualize CS1 concepts. Our results showed that professors could relate topics and concepts with 54 observations in the training set. Each cluster's main topic was explicit in the theme label task, and the clusters contained different concepts per topic. The professors can use this information to understand how often concepts cooccur while solving exercises. A future research direction is necessary to investigate if more specific clusters could be produced by increasing the number of samples or using more complex models.

Although there are potentials to use these experiments in a real-world environment, they require further research to effectively and practically integrate them into professors' working tools. These tools should support other programming languages as well.
\section*{Acknowledgment}
We want to thank all the professors who contributed to the research by providing CS1 problems and respective solutions or evaluating the results.

% \section{Bibliography styles}

% There are various bibliography styles available. You can select the style of your choice in the preamble of this document. These styles are Elsevier styles based on standard styles like Harvard and Vancouver. Please use Bib\TeX\ to generate your bibliography and include DOIs whenever available.

% Here are two sample references: \cite{Feynman1963118,Dirac1953888}.

% \section*{References}
% \newpage
\bibliography{bibliography}
% \bibliography{bibliography_correct}

% \newpage
\vskip 0pt plus -1fil
\begin{IEEEbiography}[{\includegraphics[width=1in,height=1.25in,clip,keepaspectratio]{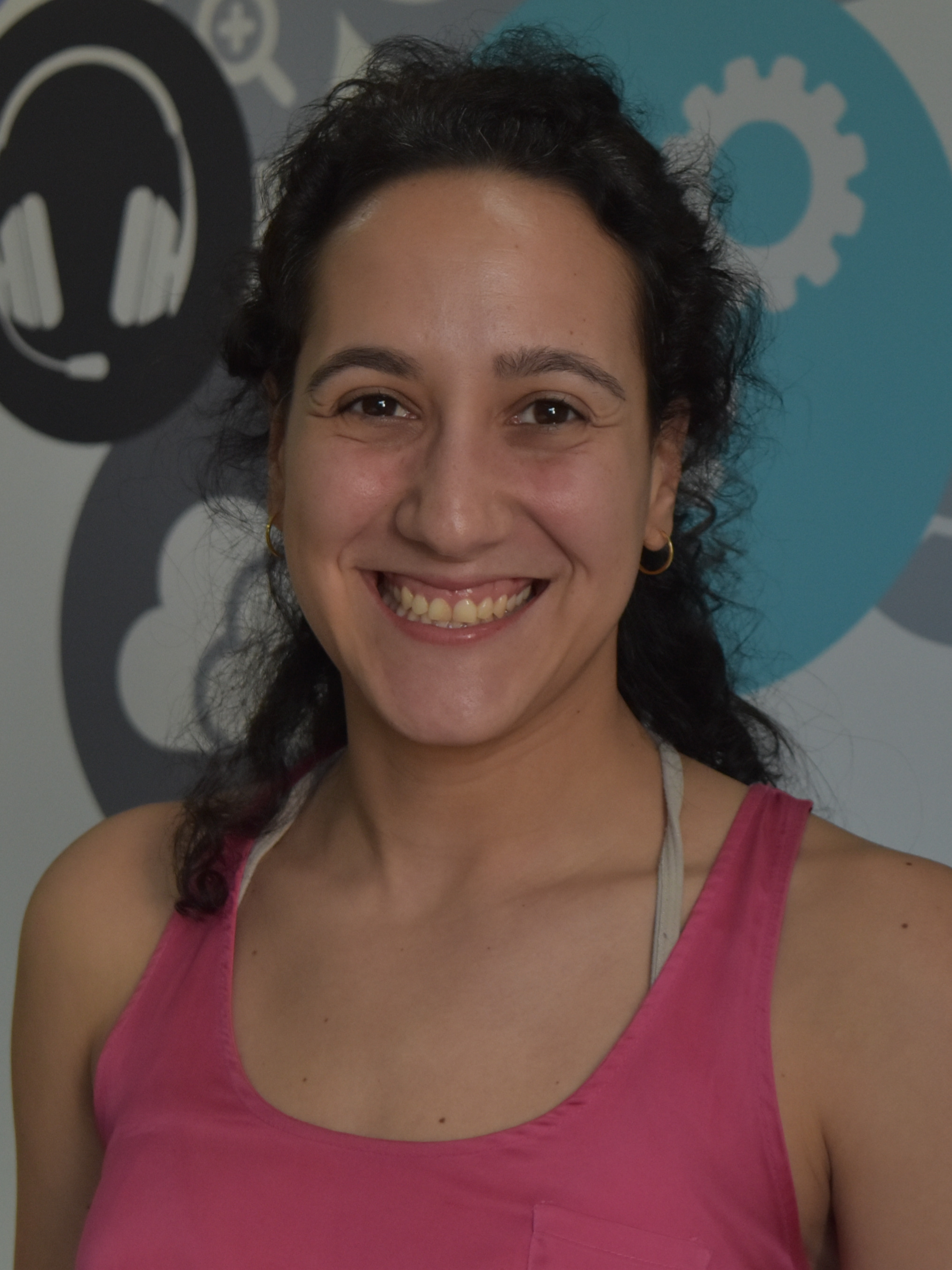}}]{Laura O. Moraes}

Laura O. Moraes received her master degree in Systems and Computing Engineering from UFRJ, Rio de Janeiro, Brazil, 2016. She is currently  pursuing  a  Ph.D. degree in the same department. In 2018, she worked as a fellow at the Data Science for Social Good Europe, organized by the University of Chicago with Nova SBE. Her research interests include text classification, item to skill mapping and the relationship between skills and student performance.
\end{IEEEbiography}
% \newpage
\vskip 0pt plus -1fil
\begin{IEEEbiography}[{\includegraphics[width=1in,height=1.25in,clip,keepaspectratio]{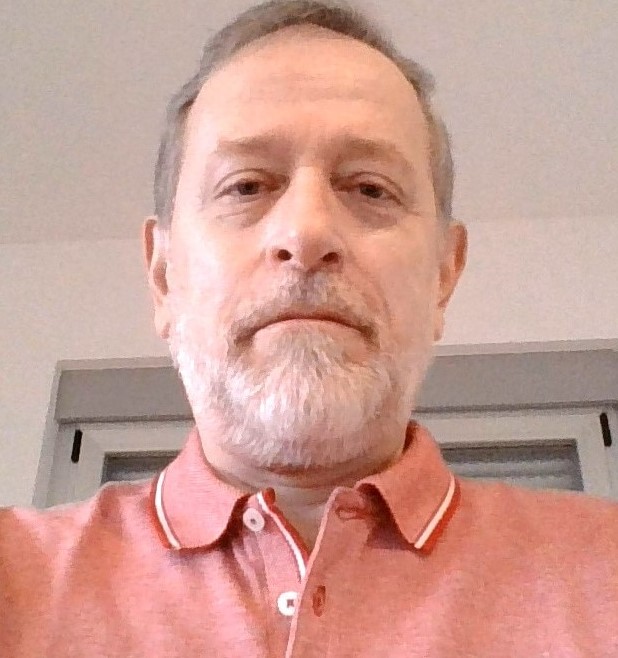}}]{Carlos Eduardo Pedreira}
Prof. Carlos Eduardo Pedreira holds Bachelor (1975) and MSc degrees
(1981) in electrical engineering from the Catholic University of Rio de Janeiro. In 1987, he received a Ph.D. degree from Imperial College of Science, Technology and Medicine, University of London. Presently, he is a Professor at the Federal University of Rio de Janeiro where he is the head of the AI sector at the Systems and Computing Department at COPPE. He is a visiting researcher at the University of Salamanca, Spain since 2002. His main research interests are Pattern Classification, Machine Learning and its applications in health. He was the founding president of the Brazilian Society of Computational Intelligence. Received the Santander Bank Award of Science and Innovation in 2006.
\end{IEEEbiography}
\end{document}